\pdfoutput=1
\documentclass[12pt]{article}

\usepackage{sbc-template}
\usepackage[utf8]{inputenc}
\usepackage[T1]{fontenc}
\usepackage[english]{babel}
\usepackage{graphicx,url}
\usepackage{amsmath,amssymb}
\usepackage{booktabs}
\usepackage{array}
\usepackage{xcolor}

\graphicspath{{figures/}}
\newcommand{\unit}[1]{\,\mathrm{#1}}

\title{A Low-Cost, Open Platform for End-to-End\\
Autonomous Driving on a Miniature Ackermann Vehicle}

\newif\ifblind
\blindfalse
\ifblind
  \author{}
  \address{}
\else
  \author{Gustavo Claudio Karl Couto\inst{1}, Eric Aislan Antonelo\inst{1}, Gabriel George Zipperer\inst{1}}
  \address{Department of Automation and Systems Engineering\\
    Federal University of Santa Catarina (UFSC) -- Florian\'opolis -- SC -- Brazil
    \email{gustavo.karl.couto@posgrad.ufsc.br, eric.antonelo@ufsc.br,}\\[-3pt]
    {\footnotesize\texttt{gabriel.zipperer@grad.ufsc.br}}}
\fi

\begin{document}

\maketitle

\begin{abstract}
This paper presents a low-cost, open experimental platform for research in end-to-end autonomous driving with miniature Ackermann vehicles. The platform combines a physical vehicle, a printed urban track, data collection tools, trajectory registration, and a Webots digital twin, enabling controlled experiments that connect simulation-based autonomous-driving methods to real-world execution. As a first baseline, we implement command-conditioned behavior cloning, in which a neural policy receives an on-board camera image and a high-level navigation command and outputs steering and speed. The system is evaluated both on the physical vehicle and in simulation. In real closed-loop experiments, the learned policy follows lanes and executes commanded turns, reaching a mean cross-track error of $6.1\unit{cm}$ with respect to the reference route, close to the $4.7\unit{cm}$ observed in human demonstrations. In the digital twin, camera field of view has a strong effect on performance, reducing the mean cross-track error from $35.6$ to $3.3\unit{cm}$ when widened from $58^\circ$ to $120^\circ$. Using the digital twin to generate synthetic driving data and a learned sim-to-real image translator to reduce the appearance gap, we further show that a higher-capacity policy trained on this synthetic data combined with real demonstrations is the only configuration that completes all four track routes in closed loop, whereas the compact baseline and the same network trained on real data alone complete fewer. These results establish the open platform as a practical testbed for sim-to-real studies and provide an initial command-conditioned imitation-learning baseline; we release it to support reproducible research.
\end{abstract}

\section{Introduction}
\label{sec:intro}

Autonomous driving systems have traditionally been organized as modular pipelines, in which perception, localization, planning, and control are designed as separate components~\cite{paden2016survey}. In parallel, learning-based approaches have explored end-to-end policies that map sensor observations directly to driving actions, reducing the amount of hand-crafted intermediate structure required by the system~\cite{pomerleau1991alvinn}. Among these methods, imitation learning is particularly attractive because it can train driving policies from demonstrations collected by a human or privileged controller.

Large-scale simulators have played an important role in this research direction. Environments such as CARLA provide controllable urban scenarios, repeatable evaluation protocols, and access to privileged information that is difficult to obtain in the real world~\cite{dosovitskiy2017carla}. They have enabled the development and comparison of end-to-end driving architectures, including policies conditioned on high-level navigation commands~\cite{codevilla2018end}, methods that learn from privileged teachers~\cite{chen2019cheating}, and approaches based on intermediate scene representations. However, simulation alone does not expose several practical constraints present in physical systems, such as camera latency, limited field of view, actuator imperfections, calibration errors, odometry drift, and imperfect synchronization between sensing and control.

Low-cost model-scale vehicles offer a useful intermediate step between simulation and full-size autonomous cars. They make it possible to evaluate autonomous-driving methods on physical hardware while preserving safety, repeatability, and accessibility~\cite{bechtel2018deeppicar,okelly2020f1tenth,hyldmar2019fleet}. Such platforms are especially valuable when they are paired with a digital twin, since the same track, vehicle geometry, and evaluation metrics can be used in both simulated and real experiments. This creates a practical setting for sim-to-real studies and for re-evaluating architectures first developed in high-fidelity simulators under real sensing and control constraints.

This paper presents a low-cost, open experimental platform for end-to-end autonomous driving with a miniature Ackermann vehicle. The platform includes a physical vehicle, an urban printed track, a data-collection procedure, a map-registration tool for quantitative trajectory evaluation, and a Webots digital twin. The goal is not only to demonstrate autonomous lane following on a small vehicle, but also to provide a controlled experimental setup in which simulation-based driving methods can be transferred, adapted, and compared in the real world.

As an initial learning baseline, we consider command-conditioned behavior cloning. A purely reactive image-to-action policy is ambiguous at intersections: the same visual observation may be compatible with turning left, turning right, or continuing straight. Conditional imitation learning addresses this ambiguity by providing a high-level command to the policy~\cite{codevilla2018end}. In our setting, the policy receives an on-board camera image and a discrete command, and outputs steering and speed for the Ackermann vehicle. This baseline is intentionally simple, serving as a reference point for future experiments with more advanced architectures and learning objectives.

The main contributions of this work are:

\begin{itemize}
    \item a low-cost, open experimental platform for autonomous-driving research with a physical Ackermann mini-vehicle, an urban printed track, data-collection tools, trajectory registration, and a Webots digital twin, released to support reproducible sim-to-real research;
    \item an experimental setup that enables sim-to-real studies and the re-evaluation, on physical hardware, of end-to-end driving architectures originally developed in simulation;
    \item a command-conditioned behavior-cloning baseline for lane following and commanded turning from an on-board camera image;
    \item a sim-to-real data pipeline that records synthetic driving data in the digital twin and maps it to the camera domain with a learned image translator, together with evidence that this synthetic augmentation is what lets a higher-capacity policy complete every track route;
    \item an initial set of ablations comparing design choices such as camera field of view, command conditioning, network capacity, and the use of sim-to-real-translated synthetic data.
\end{itemize}

Experimental results show that the platform supports closed-loop autonomous driving on the physical track and provides quantitative route-following metrics through map registration. The command-conditioned baseline follows lanes and executes operator-issued turns, reaching a mean cross-track error close to that of the human demonstrations. Comparing policies, a higher-capacity network trained with sim-to-real-translated synthetic data added to the real demonstrations completes all four routes in closed loop, while the compact baseline and the real-data-only network complete fewer. In the digital twin, the camera field of view is shown to be a critical factor for route-following performance.

The remainder of the paper is organized as follows. Section~\ref{sec:related} reviews related work. Section~\ref{sec:baseline} describes the command-conditioned behavior-cloning baseline. Section~\ref{sec:agent} presents the policy architectures. Section~\ref{sec:exp} describes the physical platform, digital twin, dataset, and experiments. Section~\ref{sec:conc} concludes the paper and discusses future work.

\section{Related Work}
\label{sec:related}

Several works exploit simulation to train and evaluate end-to-end policies for urban driving. Conditional imitation learning introduced high-level navigation commands to disambiguate visuomotor policies at intersections~\cite{codevilla2018end}, while later studies analyzed the limitations of behavior cloning under distribution shift and dataset bias~\cite{codevilla2019limitations}. Other approaches use intermediate or privileged information to improve learning, for example privileged-teacher training~\cite{chen2019cheating}. Related studies have also explored implicit behavior cloning for multimodal action distributions~\cite{florence2022ibc} and diffusion-based offline behavior cloning. These works motivate physical testbeds where architectures first studied in simulation can be re-evaluated under real sensing, latency, calibration, and actuation constraints.

Low-cost model-scale vehicles provide an intermediate step between simulation-only research and full-size autonomous cars. DeepPicar demonstrated that a small vehicle with an embedded computer can run a CNN-based end-to-end controller in real time~\cite{bechtel2018deeppicar}. F1TENTH established a widely used 1/10-scale autonomous racing platform with both hardware and virtual environments for safe and repeatable experimentation~\cite{okelly2020f1tenth}. Other miniature-car platforms have also been proposed for cooperative-driving and multi-vehicle experiments~\cite{hyldmar2019fleet}. Compared with these platforms, the present work focuses on command-conditioned urban navigation on a printed road network, with a low-cost Ackermann vehicle, a data-collection pipeline, map-referenced trajectory evaluation, and a Webots digital twin. The goal is not to replace larger platforms, but to provide an accessible setup for studying sim-to-real transfer and for testing end-to-end driving architectures previously evaluated mainly in simulation.

\section{Command-Conditioned Behavior Cloning Baseline}
\label{sec:baseline}

Given an RGB image $I_i$ from the on-board camera and a high-level command $c_i$, the policy $\pi_\theta(I_i,c_i)$ predicts a continuous steering command $\hat\delta_i$ and a signed speed command $\hat v_i$. The expert label for each frame is $a_i=(\delta_i,v_i)$, collected during teleoperation. Both outputs are bounded to $[-1,1]$ by a $\tanh$ activation and then denormalized to the physical actuator limits.

The policy is trained by weighted mean squared error,
\begin{equation}
  \mathcal{L}(\theta) = \frac{1}{N}\sum_{i=1}^{N}
  \Big[ w_{\delta}\big(\delta_i-\hat{\delta}_i\big)^2
       + w_{v}\big(v_i-\hat{v}_i\big)^2 \Big],
  \label{eq:bcloss}
\end{equation}
where $w_{\delta}=1$ and $w_v\le0.25$ weight the steering and speed terms; speed is down-weighted because route-following accuracy is dominated by steering and the closed-loop speed is capped. We use deterministic bounded regression rather than a probabilistic action model to keep the baseline simple.

\subsection{Command conditioning}
A purely reactive policy cannot choose among the valid maneuvers at an intersection~\cite{codevilla2018end}. We therefore expose the intended maneuver as a discrete command $c$ and condition the policy on it, $\pi_\theta(I,c)$. The command takes one of three values, \texttt{follow\_lane}, \texttt{turn\_left}, and \texttt{turn\_right}, encoded as a one-hot vector. During data collection, the command for each frame is derived from labelled intersection spans: a turn command is active only while the vehicle is inside the labelled intersection interval, and the command is \texttt{follow\_lane} everywhere else. At test time, the operator selects it from a direction pad, holding the turn command across the intersection, so the same visual policy can be steered through different routes.

\section{Policy Architectures}
\label{sec:agent}

\subsection{Input and output representation}
The policy input is an RGB image from the vehicle's on-board camera, resized to $160\times120$ pixels, together with the one-hot navigation command. Frames are kept in RGB channel order and scaled to $[0,1]$, with no lens undistortion applied; the higher-capacity variant instead takes $320\times240$ frames with the top $35\%$ cropped to remove the region above the road. The action space is $a\in[-1,1]^2$: steering is denormalized to $\pm45^\circ$ and speed to the configured maximum (up to $0.7\unit{m/s}$, capped at $0.15\unit{m/s}$ in the physical experiments for safety).

\subsection{Compact CNN baseline}
The main baseline policy (Fig.~\ref{fig:nets}a) is a compact CNN that maps the input image into an embedding and combines it with the command to produce the two actions. The $160\times120$ image passes through four strided convolutional blocks ($16$, $32$, $64$, $96$ channels; $5\times5$ then $3\times3$ kernels; batch normalization and ReLU) followed by global average pooling, producing an embedding vector of size $96$. This embedding is concatenated with the one-hot command and processed by a two-layer MLP that maps to the desired actions through a $\tanh$ output. The whole network has $94{,}882$ parameters.

\subsection{Architecture variants for ablation}
\begin{figure}[t]
  \centering
  \includegraphics[width=\linewidth]{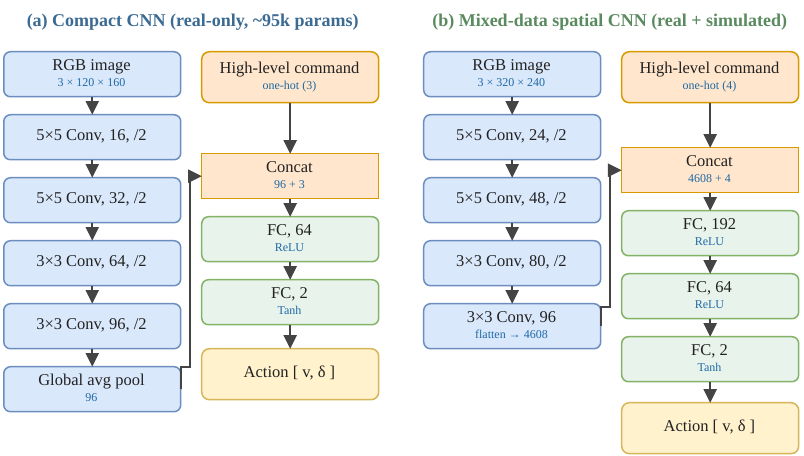}
  \caption{Policy architectures. (a) The compact command-conditioned CNN baseline
  ($94{,}882$ parameters): a convolutional encoder with global average pooling whose
  embedding is concatenated with the one-hot command and decoded by a two-layer MLP.
  (b) The larger command-conditioned variant: a spatial encoder whose flattened
  features are concatenated with the command and decoded by a three-layer MLP.}
  \label{fig:nets}
\end{figure}

For ablation we compare it with a larger command-conditioned variant (Fig.~\ref{fig:nets}b, about $1.0$M parameters) that keeps the same input-output interface and task; besides isolating the effect of capacity, the larger variant lets us compare training on real data alone against the mixed synthetic-plus-real data (Section~\ref{sec:exp}).

\section{Experiments}
\label{sec:exp}

\subsection{Vehicle platform and track}
\begin{figure}[t]
  \centering
  \includegraphics[width=\linewidth]{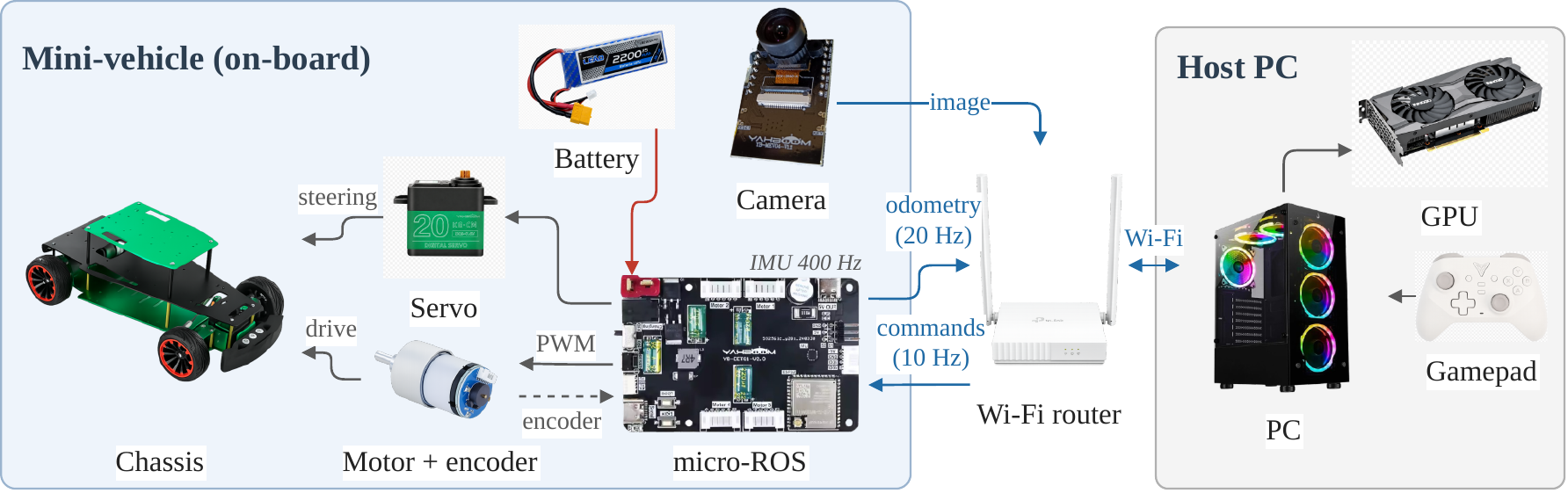}
  \caption{Experimental setup, split into the on-board mini-vehicle and the host
  PC: the ESP32-S3 micro-ROS board closes the $100\unit{Hz}$ speed PID loop and
  drives the servo, while a separate ESP32 camera streams images to the host PC,
  which runs the policy and the extended Kalman filter and returns steering and
  speed commands.}
  \label{fig:setup}
\end{figure}

The vehicle is an off-the-shelf Ackermann chassis (Yahboom ROSMASTER R2L class) with two rear drive motors, a servo-steered front axle, and a forward camera on a short mast (Fig.~\ref{fig:setup}). Its wheelbase is $0.2354\unit{m}$, its track width $0.1911\unit{m}$, and steering is limited to $\pm45^\circ$. An ESP32-S3 runs an ESP-IDF firmware with a micro-ROS client~\cite{belsare2023micro,macenski2022ros2}. It closes a per-wheel incremental-PID speed loop at $100\unit{Hz}$ (gains $0.75/0.12/0.04$, $836$-count/rev encoders), with left/right setpoints derived from the commanded speed and steering angle by the Ackermann geometry (an electronic differential), drives the steering servo, and enforces a battery cut-off and a $500\unit{ms}$ command watchdog. The firmware publishes a raw odometry twist and the IMU at $20\unit{Hz}$, which a host-side extended Kalman filter~\cite{moore2016generalized} fuses into the pose used for trajectory plots. A separate ESP32 camera streams QVGA images over a low-latency UDP link whose measured transport delay has a median of $27\unit{ms}$ and a 95th percentile of $31\unit{ms}$; the end-to-end capture-to-actuation latency (including $2.5\unit{ms}$ of policy inference) has a median of $59.5\unit{ms}$ and a 95th percentile of $118.7\unit{ms}$, with the policy running at $10\unit{Hz}$. The camera is calibrated at QVGA from a ChArUco board's chessboard corners~\cite{zhang2000flexible,garrido2014automatic}.\footnote{Firmware, host software (operator node, policy inference, camera bridge, EKF and calibration files) and the deployed checkpoint are released under the MIT license at \url{https://github.com/gustavokcouto/teledriving-in-miniature}.}

\begin{figure}[t]
  \centering
  \includegraphics[width=\linewidth]{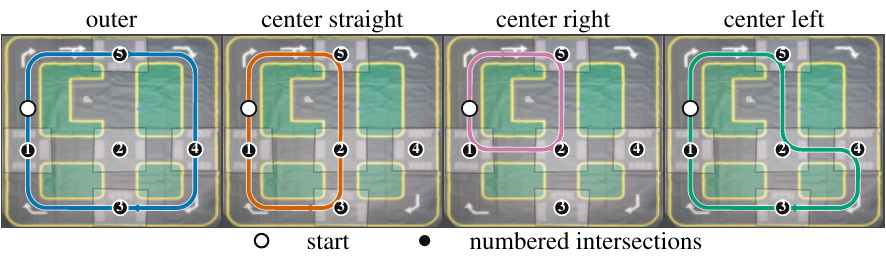}
  \caption{The four ideal routes on the rectified mat used as the Webots floor
  texture. Each panel shows one route, driven in a single direction from the
  shared start marker (white circle). Numbered markers indicate the labelled
  intersections used for command annotation.}
  \label{fig:routes}
\end{figure}

The experiments are run on a printed road mat of $3.2\times2.8\unit{m}$. All streets are driven in a single direction, which yields four selectable loops (\texttt{outer}, \texttt{center\_left}, \texttt{center\_right}, and \texttt{center\_straight}) that also define the high-level commands (Fig.~\ref{fig:routes}). To obtain a metric reference frame, the mat is photographed and perspective-rectified into a flat top-down texture. Road intersections are marked manually with a companion interface in which the operator draws, for each intersection, the four straight road edges that bound it (shown demarcated in Fig.~\ref{fig:routes}); the region they enclose defines the intersection span later used to assign the high-level command. The rectified map is reused as the Webots floor texture and as the common reference frame for labelled intersections, ideal routes, and cross-track error.

\subsection{Dataset generation}
\begin{figure}[t]
  \centering
  \includegraphics[width=333pt]{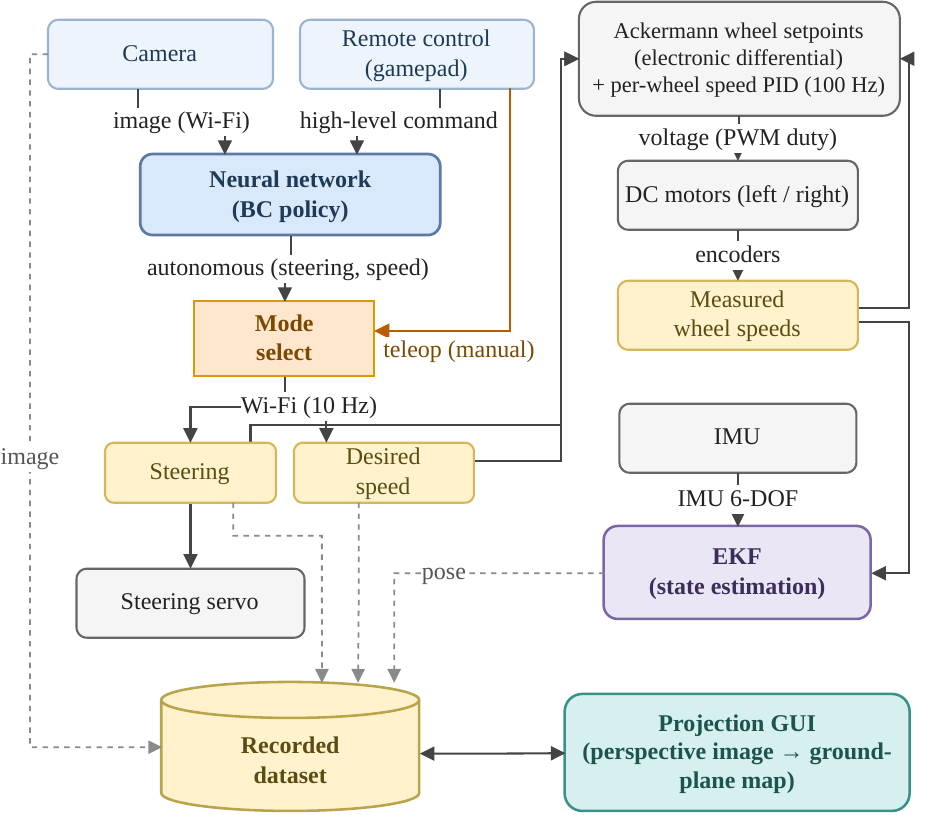}
  \caption{End-to-end signal flow. During data collection a human teleoperates the
  vehicle and the mode-select logs the commanded steering/speed---with the camera
  image and the EKF pose---into the recorded dataset; at deployment the
  behavior-cloning policy replaces the teleoperator, producing steering and a speed
  setpoint; on board, steering drives the servo and, with the speed setpoint,
  defines per-wheel setpoints (Ackermann electronic differential) for the
  $100\unit{Hz}$ PID loops, while the projection GUI maps recorded images to the
  ground plane.}
  \label{fig:control}
\end{figure}

The expert dataset is generated by a human driver who teleoperates the vehicle with a gamepad while holding a deadman trigger that gates the recording; Fig.~\ref{fig:control} summarizes the resulting end-to-end signal flow, from this manual collection to autonomous deployment. Each session is stored as a folder of $320\times240$ frames recorded at $10\unit{Hz}$ together with labels for steering, speed, fused odometry pose, IMU, encoder signals, and command gate. The expert dataset comprises four route families with $41$ sessions and $19{,}206$ frames, of which $18{,}467$ are active driving frames ($5{,}137$ for \texttt{center\_left}, $3{,}544$ for \texttt{center\_right}, $4{,}512$ for \texttt{center\_straight}, and $5{,}274$ for \texttt{outer}). High-level commands are assigned from labelled intersection spans ($75$ right-turn, $10$ left-turn, and $78$ straight segments), remapping the recorded straight action to \texttt{follow\_lane}.

\subsection{Synthetic data and sim-to-real translation}
The higher-capacity policy is additionally trained with synthetic data from a Webots digital twin of the same track. A scripted controller follows the ideal routes of the road network and records clean expert steering and speed labels at $10\unit{Hz}$, while actuator-noise and off-route recovery perturbations enrich the visited states without altering those expert labels. This produces $96$ laps ($24$ for each of the four routes) and $43{,}581$ active frames.

To narrow the appearance gap between the simulator and the on-board camera, each synthetic frame is converted offline by a sim-to-real image translator before training; only the image is changed, while the expert labels are kept. The translator is a four-level U-Net~\cite{ronneberger2015unet} trained with a paired $L_1$ objective on $2{,}559$ registered simulator--real image pairs (from $16$ real sessions), applying photometric domain randomization~\cite{tobin2017domain} to the synthetic input. Among the methods we compared---this domain-randomized U-Net, pix2pix~\cite{isola2017pix2pix}, and SimGAN~\cite{shrivastava2017simgan}---it produced the lowest closed-loop route error in the twin, reducing the mean route error of a real-trained policy from $28.0$ to $13.1\unit{cm}$. The translated synthetic laps are then concatenated with the $41$ real sessions, giving a mixed dataset of $137$ sessions and $62{,}048$ active frames.

\subsection{Training protocol}
Policies are trained by behavior cloning with the weighted loss of Equation~\eqref{eq:bcloss} on the active frames, using AdamW (learning rate $10^{-3}$, weight decay $10^{-4}$) and batch size $32$, holding out whole sessions for validation to avoid leakage between adjacent frames. The compact baseline is trained on the real dataset, whereas the higher-capacity policy is trained either on the real dataset alone or on the mixed synthetic-plus-real dataset, with speed- and steering-loss weights of $0.25$ and $1.0$. For the mixed dataset, the policy is trained for $200$ epochs and the checkpoint is selected by closed-loop rollout in the digital twin rather than by lowest validation loss, since the two criteria do not coincide.

\subsection{Map registration and route-deviation metric}
\begin{figure}[t]
  \centering
  \includegraphics[width=0.88\linewidth]{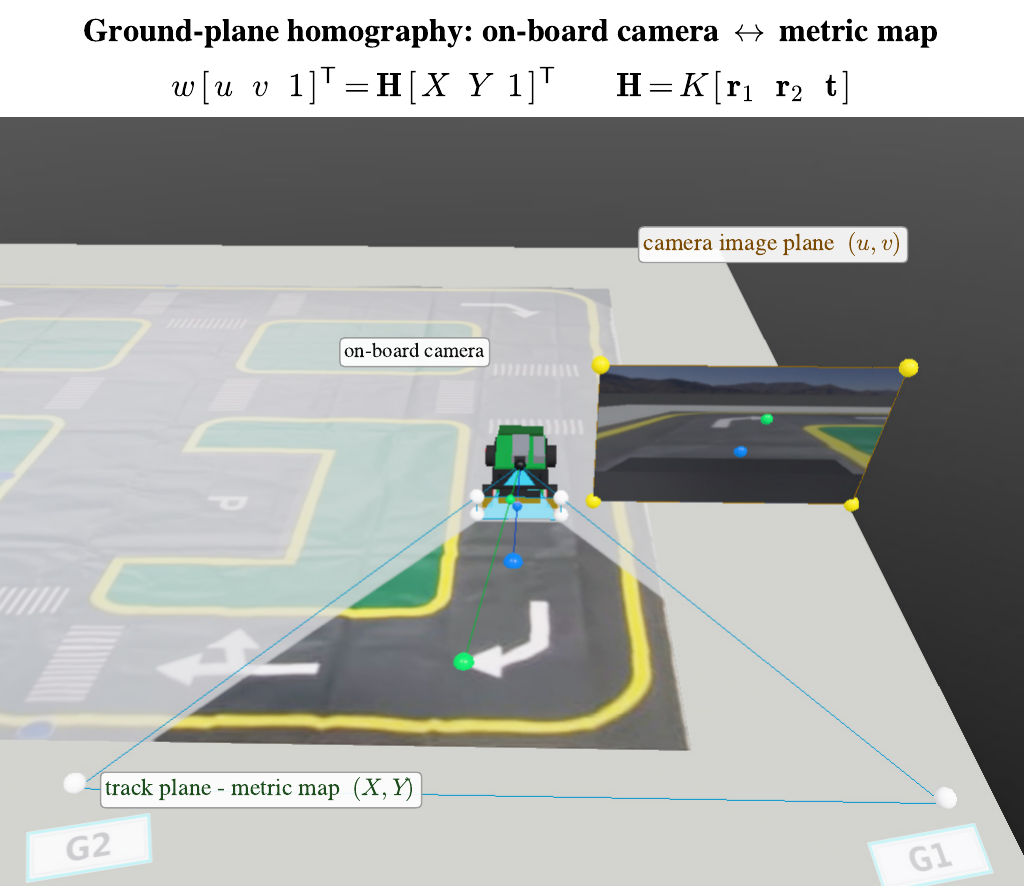}
\caption{Ground-plane homography. Because the track is planar, camera intrinsics
and pose define a single $3\times3$ homography between metric map coordinates
$(X,Y)$ and image pixels $(u,v)$. The projection tool uses this mapping, or its
inverse, to relate on-board camera observations to the rectified map; the figure
illustrates the geometry in the Webots digital twin.
The blue and green markers are the same
physical points on the track and in the camera image.}
  \label{fig:homography}
\end{figure}

\begin{figure}[t]
  \centering
  \includegraphics[width=0.86\linewidth]{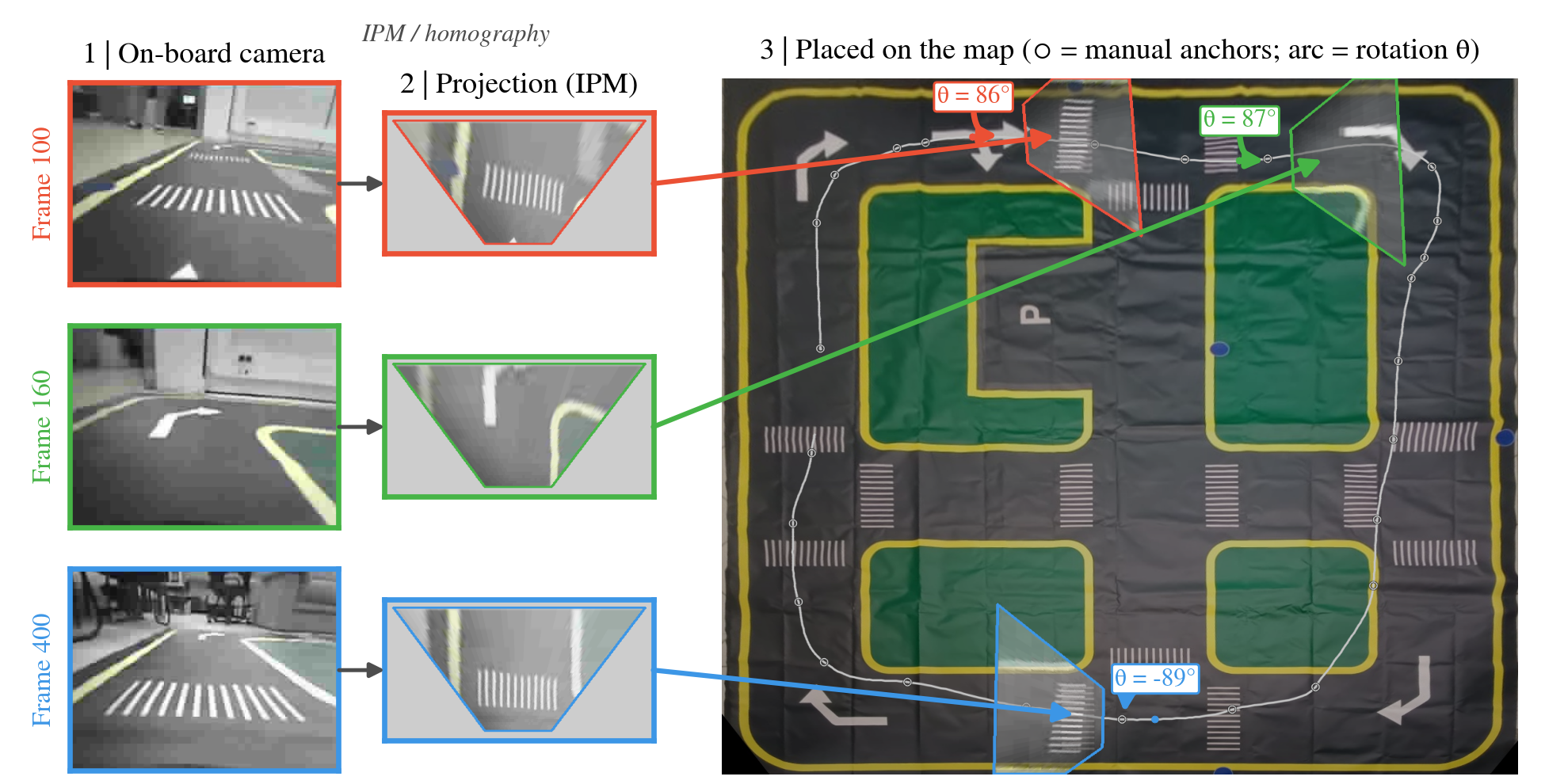}
  \caption{Map-registered localization of an outer-loop run: per-frame camera images
  are projected to the ground plane and the reconstructed trajectory is overlaid on
  the rectified map, illustrating the registration behind the cross-track-error
  metric.}
  \label{fig:projmulti}
\end{figure}

The dataset images are local camera observations, whereas the route is defined on the global map. To obtain a map-referenced accuracy metric, each run is registered to the rectified track with a calibrated tool that projects the camera image onto the ground plane using intrinsics and extrinsics (the planar homography of Fig.~\ref{fig:homography}), and an operator aligns the projected trajectory to the map. This yields a globally consistent trajectory $p_i=(x_i,y_i)$ in metres (Fig.~\ref{fig:projmulti}). The ideal route centre lines are built from the track loops and the marked intersections. For a registered point $p_i$ and the ideal polyline $\Gamma_r$ of route $r$, the cross-track error is
\begin{equation}
  e_i = \min_{q\in\Gamma_r}\lVert p_i-q\rVert_2 .
  \label{eq:xtrack}
\end{equation}
We report per-run mean, median, and 95th-percentile errors, scoring only fully localized runs---those with at least ten manual anchors spanning $\ge85\%$ of their frames---namely $10$ human sessions and the three neural-control runs.

\subsection{Physical closed-loop evaluation}
We evaluate three fully autonomous runs of the compact baseline (about $1{,}300$ frames each), in which the agent drives continuously while the operator only issues high-level commands and the speed is capped at $0.15\unit{m/s}$. Each run repeats the outer, center-straight, and center-right loops; the more demanding center-left (left-turn) loop is completed only by the larger mixed-data policy (Fig.~\ref{fig:policy_ablation_laps}). Steering stays centred under \texttt{follow\_lane} and consistently negative under \texttt{turn\_right} (median around $-22^\circ$), with no operator interventions. Registering each run to the map (Eq.~\eqref{eq:xtrack}) and matching every lap to its ideal loop, the policy drives with a $6.1\pm1.2\unit{cm}$ mean cross-track error, close to the $4.7\pm1.8\unit{cm}$ of the fully localized human demonstrations---a reference for the available demonstrations rather than a strict upper bound, since teleoperation itself introduces some tracking error.

\subsection{Digital twin and camera field-of-view ablation}
The physical metric conflates many factors. To isolate camera field of view with exact ground truth, we evaluate the same policy in the Webots~\cite{michel2004webots} digital twin (the same track and the vehicle's Ackermann geometry---wheelbase, track width, and wheel sizes) while varying the simulated lens. The calibrated narrow lens ($58^\circ$) follows the lane only loosely, whereas widening it to $120^\circ$ cuts the mean cross-track error from $35.6\pm12.1\unit{cm}$ (per-route $25$--$53\unit{cm}$, with peaks above $1\unit{m}$) to $3.3\pm0.2\unit{cm}$ (per-route $3.1$--$3.5\unit{cm}$), identifying field of view as a major limiting factor.

\subsection{Architecture and command-conditioning ablations}
\begin{figure}[ht]
  \centering
  \includegraphics[width=0.76\linewidth]{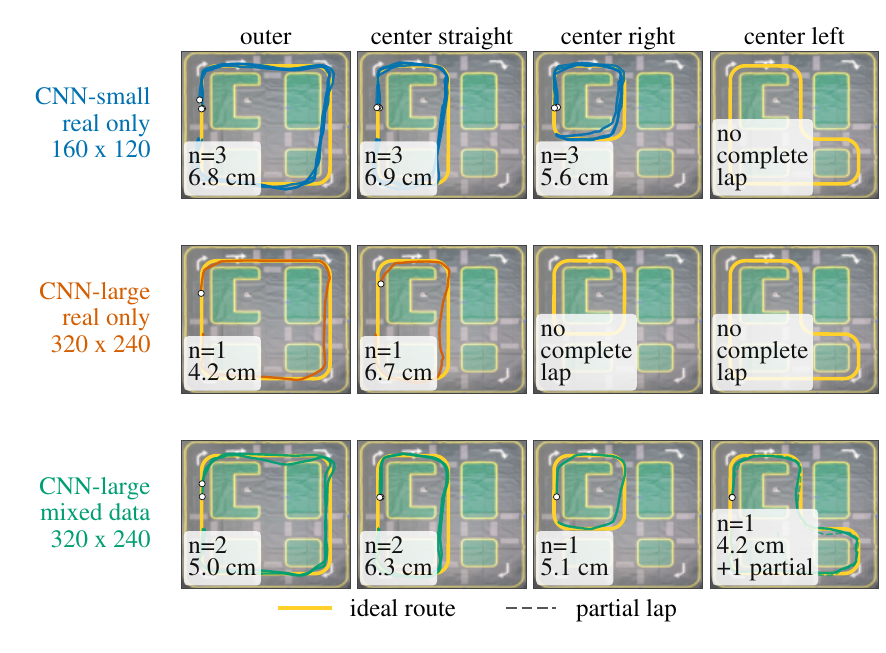}
  \caption{Comparison of map-registered neural-control laps by policy variant. Each row groups the autonomous runs from one policy; columns show the four route loops. Colored lines are registered real-robot laps, dashed lines are partial laps, yellow lines are ideal routes, and each panel reports the mean cross-track error over complete laps.}
  \label{fig:policy_ablation_laps}
\end{figure}

Keeping the command input fixed, we compare the compact baseline (CNN-small) with the larger policy (CNN-large) under two training sets, real-only and mixed synthetic-plus-real. The clearest difference is how many of the four routes each policy completes in closed loop (Fig.~\ref{fig:policy_ablation_laps}). The compact baseline completes the outer, center-straight, and center-right loops but fails the center-left turn; the larger policy trained on real data alone completes only the outer and center-straight loops; and the larger policy trained on the mixed dataset is the only one to complete all four routes. On the routes it completes, the mixed-data policy stays close to the ideal path, with per-route mean cross-track errors of $3.9$--$6.1\unit{cm}$ (outer), $6.0$--$6.5\unit{cm}$ (center-straight), $5.1\unit{cm}$ (center-right) and $4.2\unit{cm}$ (center-left), for an overall mean of about $5.1\unit{cm}$. The added capacity thus generalizes to every maneuver only when the sim-to-real-translated synthetic data broadens the training distribution.

\section{Conclusion}
\label{sec:conc}

This work presented a low-cost, open sim-to-real platform for end-to-end autonomous driving with an Ackermann mini-vehicle, together with a command-conditioned behavior-cloning baseline. On the physical track, the compact policy follows lanes and executes operator-issued turns at a $6.1\unit{cm}$ mean cross-track error, close to the $4.7\unit{cm}$ of the human demonstrations; a higher-capacity network completes all four routes only when its real data is augmented with sim-to-real-translated synthetic laps; and in the digital twin, widening the camera from $58^\circ$ to $120^\circ$ cuts cross-track error from $35.6$ to $3.3\unit{cm}$.

The main limitations are the small number of physical runs, the manual component of the trajectory registration, and the preliminary ablations, which lack a compact-network/mixed-data condition and confound capacity with input resolution. Future work includes automatic image-based registration, a wider-field-of-view camera, and stronger baselines (BEV-based, multimodal, diffusion or transformer policies).

\section*{Acknowledgment}
This work was partially funded by the National Council for Scientific and Technological Development -- CNPq, Brazil (Grant No.\ 420148/2025-6), and by the Coordena\c{c}\~ao de Aperfei\c{c}oamento de Pessoal de N\'ivel Superior -- Brasil (CAPES) -- Finance Code 001. G.~G.~Z. also acknowledges CNPq for the undergraduate research scholarship granted through PIBIC.

\bibliographystyle{sbc}
\bibliography{references}

\end{document}